\documentclass[10pt,twocolumn,letterpaper]{article}
\usepackage[pagenumbers]{cvpr} 

\usepackage{graphicx}
\usepackage{amsmath}
\usepackage{amssymb}
\usepackage{booktabs}
\usepackage{colortbl}
\usepackage{booktabs}   
\usepackage{epsfig}
\usepackage{tabularx}
\usepackage{textcomp}      	
\usepackage{wrapfig}
\usepackage{multirow}
\usepackage[table]{xcolor}
\usepackage{adjustbox}
\usepackage{makecell}
\usepackage{amsmath}
\usepackage{mathtools}
\usepackage[accsupp]{axessibility}

\DeclarePairedDelimiter\ceil{\lceil}{\rceil}

\usepackage[ruled,vlined]{algorithm2e}

\SetCommentSty{mycommfont}
\newcommand{\calD}{{\mathcal{D}}}
\newcommand{\calE}{{\mathcal{E}}}

\newcommand{\calL}{{\mathcal{L}}}

\newcommand{\be}{\begin{eqnarray}}
\newcommand{\ee}{\end{eqnarray}}
\newcommand{\bee}{\begin{eqnarray*}}
\newcommand{\eee}{\end{eqnarray*}}

\newcommand{\matrixb}{\left[ \begin{array}}
\newcommand{\matrixe}{\end{array} \right]}   

\newcommand{\Tref}[1]{Table~\ref{#1}}
\newcommand{\Eref}[1]{Eq.~(\ref{#1})}
\newcommand{\Fref}[1]{Fig.~\ref{#1}}

\newcommand{\Cref}[1]{Chap.~\ref{#1}}
\newcommand{\Sref}[1]{Sec.~\ref{#1}}
\newcommand{\Aref}[1]{Algo.~\ref{#1}}

\def\eg{\emph{e.g.}}

\def\etal{\emph{et al. }}
\def\ie{\emph{i.e. }}

\usepackage[pagebackref,breaklinks,colorlinks]{hyperref}

\usepackage[capitalize]{cleveref}
\crefname{section}{Sec.}{Secs.}
\Crefname{section}{Section}{Sections}
\Crefname{table}{Table}{Tables}
\crefname{table}{Tab.}{Tabs.}

\def\cvprPaperID{2253} 
\def\confName{CVPR}
\def\confYear{2022}

\begin{document}

\title{Long-term Video Frame Interpolation via Feature Propagation}

\author{Dawit Mureja Argaw \hspace{7mm} In So Kweon\\
Korea Advanced Institute of Science and Technology\\
Daejeon, Republic of Korea\\
{\tt\small \{dawitmureja, iskweon77\}@kaist.ac.kr}
}
\maketitle

\begin{abstract}
Video frame interpolation (VFI) works generally predict intermediate frame(s) by first estimating the motion between inputs and then warping the inputs to the target time with the estimated motion. This approach, however, is not optimal when the temporal distance between the input sequence increases as existing motion estimation modules cannot effectively handle large motions. Hence, VFI works perform well for small frame gaps and perform poorly as the frame gap increases. In this work, we propose a novel framework to address this problem. We argue that when there is a large gap between inputs, instead of estimating imprecise motion that will eventually lead to inaccurate interpolation, we can safely propagate from one side of the input up to a reliable time frame using the other input as a reference. Then, the rest of the intermediate frames can be interpolated using standard approaches as the temporal gap is now narrowed. To this end, we propose a propagation network (\textsf{PNet}) by extending the classic feature-level forecasting with a novel motion-to-feature approach. To be thorough, we adopt a simple interpolation model along with \textsf{PNet} as our full model and design a simple procedure to train the full model in an end-to-end manner. Experimental results on several benchmark datasets confirm the effectiveness of our method for long-term VFI compared to state-of-the-art approaches.
\end{abstract}
\vspace{-7mm}
\section{Introduction}
\vspace{-1mm}
Video frame interpolation (VFI) aims at predicting one or more intermediate frames from a given frame sequence. Given inputs $\langle x_t$, $x_{t+n}\rangle$, where $n$ is the frame gap between the inputs, existing VFI works generally follow two steps. First, they estimate the motion between $x_t$ and $x_{t+n}$ using off-the-shelf motion estimation modules or by imposing motion constraints. Then, they warp the inputs to the target time and synthesize an intermediate frame. 

VFI works target temporal super-resolution on a premise that the frame rate of the input sequence is often already sufficiently high. We have experimentally verified this notion by evaluating several state-of-the-art VFI methods~\cite{niklaus2017video2,slomo,dain2019,Lee_2020_CVPR,gui2020featureflow} on input sequences sampled at different frame rates. Even though a reasonable performance decrease is an expected phenomenon, we observed a significant drop in performance when the frame rate of the input sequence decreases (see \Tref{tbl:temp_vfi}), highlighting that interpolating frames becomes very challenging as the temporal distance between consecutive frames increases. Moreover, far less attention has been given to this problem in past literature as most evaluations have been done on videos with fixed frame rate (mostly 30 fps). We argue that the main reason behind this limitation is partly associated with the working principle of VFI works. If the estimated motion between inputs is inaccurate, then the interpolated frame synthesized by time warping the inputs with the estimated motion will also likely be inaccurate. This is particularly problematic when the temporal gap between input frames is large as existing flow or kernel based motion estimation modules can not effectively handle large motions.

In this work, we tackle the long-term video interpolation problem and propose a general VFI framework robust to relatively low frame rates. Specifically, when there is a large gap between input frames, instead of predicting the motion between the inputs which will likely be imprecise and eventually lead to inaccurate interpolation, we conjecture that we can safely propagate from one side of the input to a reliable extent of time using the other input as a useful reference, \ie given $\langle x_t$, $x_{t+n}\rangle$, we propagate up to $x_{t+\Delta t}$ from the side of the first input $x_t$ and we similarly propagate up to $x_{t+n-\Delta t}$ from the side of the second input $x_{t+n}$, where $\Delta t$ is the extent of propagation. This is intuitive because the intermediate frames in the neighborhood of $x_t$ will most likely depend on $x_t$ compared to $x_{t+n}$, and vice versa. Once we propagate to a reliable time frame from both sides, the rest of the intermediate frames between $x_{t+\Delta t}$ and $x_{t+n-\Delta t}$  can be interpolated using existing interpolation approaches as the temporal gap is now reduced to $n-2\Delta t$. 

To this end, we propose a propagation network (\textsf{PNet}) that predicts future frames by relying more on one of the inputs while attending the other. We accomplish this by extending the classic feature-to-feature (F2F) forecasting \cite{vondrick2016anticipating,couprie2018joint,vora2018future,chiu2020segmenting,vsaric2019single,saric2020warp} with a novel motion-to-feature (M2F) approach, where we introduce optical flow as another modality to guide the propagation of features and to enforce temporal consistency between the propagated features. Unlike feature supervision which makes a network more dependent on the semantics of input frames, our motion supervision allows the network to focus on the motion between inputs and ensures features are propagated accordingly irrespective of the contents of the images. Moreover, while most F2F works focus on predicting task-specific outputs such as segmentation maps, we perform RGB forecasting by designing a frame synthesis network that reconstructs frames from the propagated features in a coarse-to-fine manner. 

We experimentally show that the proposed \textsf{PNet} can be used as a plug in module to make existing state-of-the-art VFI approaches \cite{niklaus2017video2,slomo,dain2019,Lee_2020_CVPR,gui2020featureflow} robust particularly when there is a considerable temporal gap between inputs. To be thorough, we adopt a light version of SloMo \cite{slomo} along with \textsf{PNet} as our full model and devise a simple, yet effective,  procedure to successfully train the full model in an end-to-end manner. We comprehensively analyze our work and previous methods on several widely used datasets \cite{su2017deep,Nah_2017_CVPR,galoogahi2017need} and confirm the favorability of our approach. Moreover, we carry out ablation experiments to shed light on the network design and loss function choices. 

\vspace{-2.3mm}
\section{Related Works}
\label{sec:related_works}
\vspace{-1mm}
\paragraph{Video Frame Interpolation.} Early conventional methods~\cite{zitnick2004high,mahajan2009moving} relied on the optical flow between inputs and the given image formation model for synthesizing intermediate frames. Recently, several deep network based VFI approaches have been proposed. While some works \cite{long2016learning,choi2020channel} directly predict intermediate frames, most existing approaches embed motion estimation modules in their framework. According to the type of the motion estimation module used, VFI works can be broadly categorized as: phase-based, kernel-based, flow-based and a mix of the last two. Early phase-based works~\cite{meyer2015phase,meyer2018phasenet} formulated the temporal change between the inputs as phase shifts. On the other hand, kernel-based methods such as AdaConv~\cite{niklaus2017video1} and SepConv~\cite{niklaus2017video2} estimate spatially adaptive 2D and 1D kernels, respectively. Meanwhile, due to the significant progress achieved in optical flow estimation research, flow-based interpolation approaches~\cite{slomo,liu2017video,reda2019unsupervised,son2020aim,xue2019video,niklaus2018context,park2020bmbc,Peleg_2019_CVPR,yuan2019zoom,niklaus2020softmax} have grown to be popular. DVF \cite{liu2017video} and SloMo \cite{slomo} estimated flows between input frames and directly warped them to the target intermediate time while~\cite{xue2019video,niklaus2018context,park2020bmbc,niklaus2020softmax} used a trainable frame synthesis network on the warped frames to predict the intermediate frame.

\begin{figure}[!t]
    \centering
    \includegraphics[width=1.0\linewidth,trim={21.8cm 18.9cm 28.3cm 11.7cm},clip]{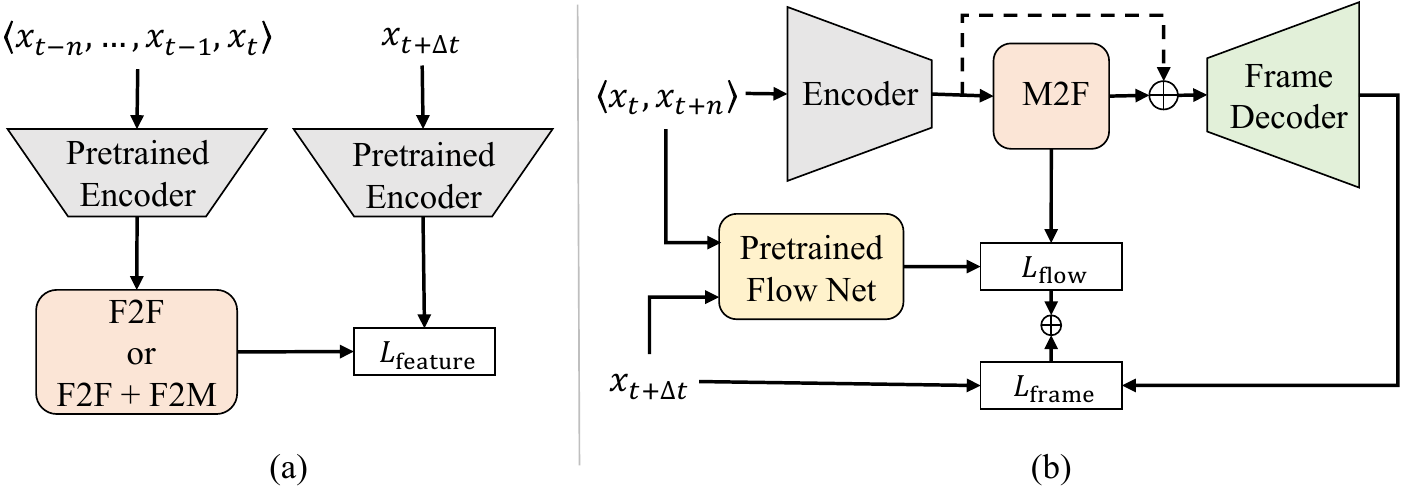}
    \vspace{-5mm}
    \caption{(a) Overview of previous feature-level propagation formulation, (b) Proposed problem formulation.}
    \label{fig:framework}
    \vspace{-5mm}
\end{figure}

DAIN~\cite{dain2019} and MEMC-Net~\cite{bao2019memc} combined kernel-based and flow-based models. AdaCoF~\cite{Lee_2020_CVPR} proposed a generalized warping module via adaptive collaboration of flows for VFI. Recently, several works~\cite{reda2019unsupervised, Peleg_2019_CVPR,choi2020scene,xu2019quadratic,chi2020all,zhangflexible,gui2020featureflow,shen2020blurry,argaw2021motion} have focused on addressing the different limitations of the VFI approaches discussed thus far. However, most existing works assume that the input sequence frame rate is often already sufficiently high, hence, long-term VFI has received far less attention in the past literature. Our work tackles this problem by proposing a novel framework that combines frame propagation and interpolation.
\vspace{-3mm}
\paragraph{Feature Propagation.} Feature-to-feature (F2F) forecasting inputs intermediate features of the past frames and anticipates their future counter parts. This approach have been previously used for action recognition \cite{vondrick2016anticipating}, instance segmentation \cite{couprie2018joint,luc2018predicting,sun2019predicting} and semantic segmentation \cite{vora2018future,chiu2020segmenting,vsaric2019single,saric2020warp} tasks. Recently, \v{S}ari\'{c} \etal \cite{saric2020warp} proposed a feature-to-motion (F2M) module to compliment the classic F2F approach. Previous F2F or F2F + F2M based works use the encoder part of task-oriented pretrained models (\eg~semantic segmentation) to extract intermediate features of a set of inputs and use the extracted features to forecast features of future frame. The forecasting module is trained by optimizing the loss between the forecasted features and the extracted features of the future frame (see \Fref{fig:framework}a). During inference, the task-specific output (\eg~segmentation map) is obtained by feeding the forecasted feature into the decoder part of the pretrained model. 

In this work, we extend feature-level propagation to a relatively unexplored task, \ie long-term VFI, by presenting a novel motion-to-feature (M2F) approach. Our approach is different from previous feature-level propagation methods in the following aspects. First, we introduce motion (in the form of optical flow) as another modality to guide the forecasting of features and to enforce temporal consistency between forecasted features. Second, we perform RGB forecasting by designing a frame synthesis network that outputs future frames from forecasted features. The proposed formulation is summarized in \Fref{fig:framework}b.

\begin{figure*}[!t]
    \centering
    \includegraphics[width=1.0\linewidth,trim={3.9cm 2.8cm 6.9cm 6.65cm},clip]{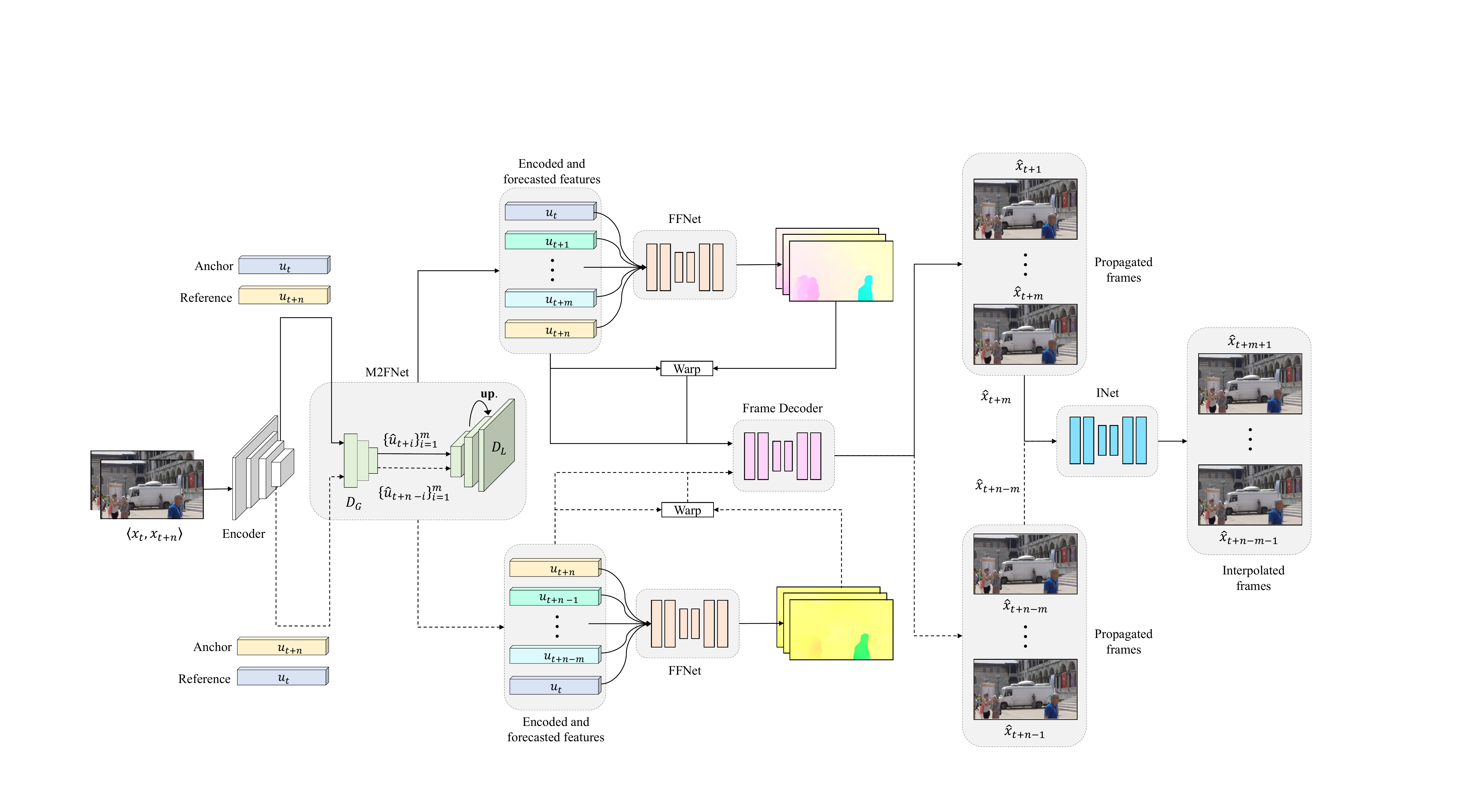}
    \vspace{-6mm}
    \caption{Overview of the proposed propagation-interpolation network (\textsf{P-INet)}. The propagation network (\textsf{PNet}) consists of an encoder network for feature extraction, \textsf{M2FNet} to bidirectionally propagate features using the encoded features as anchor and reference features, \textsf{FFNet} to estimate optical flow between features for motion supervision, and a frame decoder to reconstruct frames from propagated features. An interpolation network (\textsf{INet}) is used to interpolate the intermediate frames between the end propagated frames.}
    \vspace{-5.5mm}
    \label{fig:model_new}
\end{figure*}
\vspace{-1mm}
\section{Methodology}
\label{sec:methodology}
Given a pair of consecutive frames $\langle x_t,x_{t+n}\rangle$ from a low-frame-rate video, we aim to generate a high quality, high-frame-rate sequence $\{x_t, x_{t+1}, \ldots, x_{t+n-1}, x_{t+n}\}$ by jointly optimizing interlinked propagation and interpolation networks in an end-to-end manner. The overview of our proposed framework is shown in \Fref{fig:model_new}. 

\subsection{Propagation Network (\textsf{\textbf{PNet}})}
\label{sec:pnet}
\begin{figure}[!t]
    \centering
    \includegraphics[width=1.0\linewidth,trim={25.5cm 22.6cm 25.25cm 6.3cm},clip]{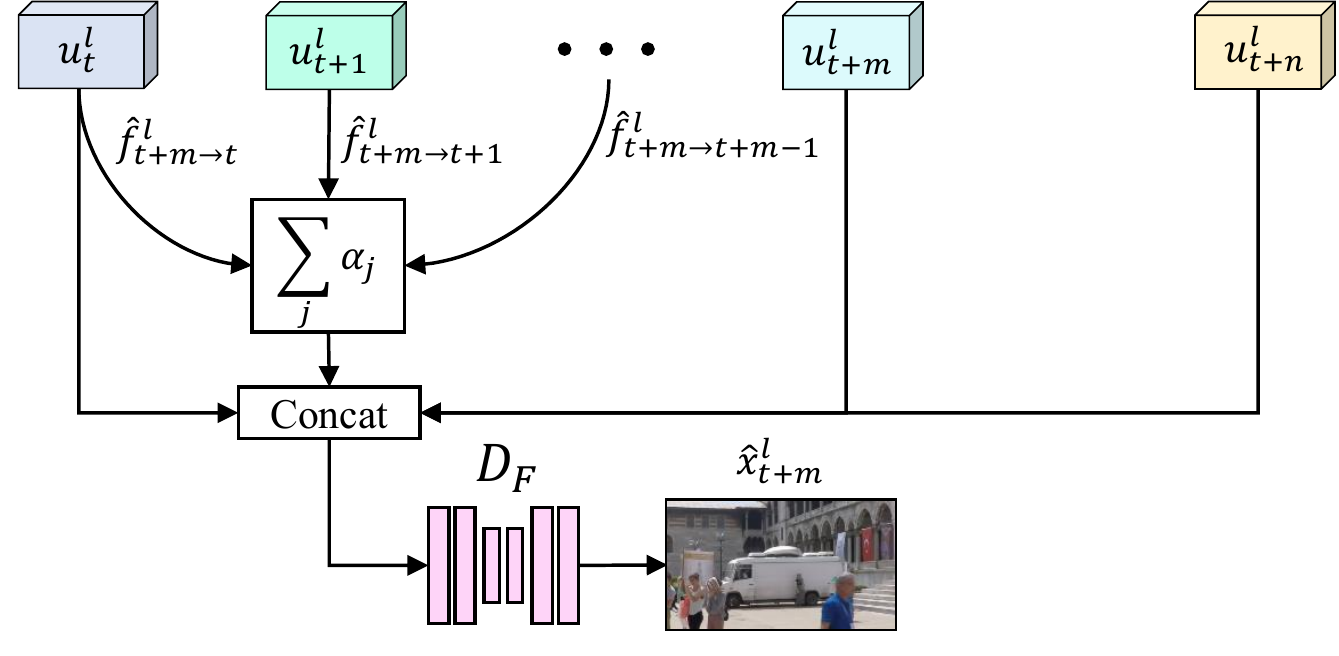}
    \caption{Frame synthesis of $\hat{x}_{t+m}$ at feature level $l$.}
    \label{fig:frame_synthesis}
    \vspace{-5.6mm}
\end{figure}

\vspace{-0.5mm}
We use an encoder-decoder architecture for \textsf{PNet}. First, we design an encoder network $\calE$ to extract features from the input frames $\langle x_t,x_{t+n}\rangle$ in a top-down manner (\Eref{eq: encode}). The encoder is a feed-forward network with 5 convolutional blocks, each block containing 2 layers of convolution with kernel size $3 \times 3$. Except for the first block, features are downsampled to half of their spatial size and the number of channels is doubled after each convolutional block.
\begin{equation}
    \big\{u_t^l\big\}_{l=1}^{k} = \calE (x_t)\hspace{4mm}\big\{u_{t+n}^l\big\}_{l=1}^{k} = \calE (x_{t+n})
    \label{eq: encode}
\end{equation}
where $l$ denotes a level in the feature pyramid with a total of $k$ levels ($k = 5$ in our experiments) and $u_t^l$ denotes an encoded feature of the first input $x_t$ at level $l$.  To propagate to the frame $x_{t + \Delta t}$ (from $x_t$ side), we first perform feature-level forecasting using the encoded features of $x_t$ and $x_{t+n}$, \ie $\{u_t^1, \ldots, u_t^k\}$ and $\{u_{t+n}^1, \ldots, u_{t+n}^k\}$, as \textit{anchor} and \textit{reference} features, respectively. We then use a decoder network to reconstruct $x_{t + \Delta t}$ from the propagated features in a bottom-up manner.
\vspace{-3.4mm}
\paragraph{Motion-to-Feature Forecasting.} We design a motion-to-feature network (\textsf{M2FNet}) to forecast the future counterparts of the encoded features. \textsf{M2FNet} takes the anchor and reference features as inputs and  anticipates the motion to propagate the anchor feature to its future counterparts. Then, it transforms the anchor feature  according to the estimated motion. To take the complex motion dynamics between the input frames into account and better exploit the inter-frame spatio-temporal correlation, we propagate to multiple frames simultaneously. \textsf{M2FNet} has 2 components: global ($\calD_G$) and local ($\calD_L$) motion decoders. $\calD_G$ learns the global motion between the encoded features, and predicts affine transformation parameters $\theta_{[R|T]}$ to spatially transform the anchor feature to its future counterparts (see \Eref{eqn:global1} and \Eref{eqn:global2}). We use spatial transformer network~\cite{jaderberg2015spatial} for for $\calD_G$. 
\vspace{-1.5mm}
\begin{equation}
  \big\{\theta_{[R_{t+i}|T_{t+i}]}\big\}_{i=1}^{m} = \calD^l_{G}\big(u_t^l~||~u^l_{t+n}\big)
    \label{eqn:global1}
\vspace{-3mm}
\end{equation}
\vspace{-3mm}
\begin{equation}
    \big\{\hat{u}^l_{t+i}\big\}_{i=1}^{m} = \textsf{transform}\ \big(u_t^l,\ \big\{\theta_{[R_{t+i}|T_{t+i}]}\big\}_{i=1}^{m}\big)
    \label{eqn:global2}
\end{equation}
where $||$ denotes channel-wise concatenation, $m$ refers to the number of features (frames) propagated from the anchor feature $u_t$ and $\hat{u}^l_{t+i}$ represents the output of $\calD_G$ at time step $t+i$ and feature level $l$. As $\calD_G$ is limited to learning only non-local motion, in order to capture the locally varying motion, we further refine the outputs of $\calD_G$ with a local motion decoder ($\calD_L$). $\calD_L$ inputs the globally transformed feature $\hat{u}_{t+i}$ along with the anchor and reference features, and outputs the forecasted feature $u_{t+i}$ (see \Eref{eqn:local}). $\calD_L$ has 3 densely connected convolutional layers \cite{huang2017densely} each with kernel size $3 \times 3$ and stride $1$. As the forecasted feature $u_{t+i}$ is decoded in a coarse-to-fine manner, a residual connection is built by feeding the the upsampled decoded feature from previous feature level $l+1$ into $\calD_L$ as shown in \Eref{eqn:local}. A deconvolution layer of kernel size $4 \times 4$ and stride size $2$ is used to upsample ($\times 2$) features.
\begin{equation}
    u_{t+i}^l= \calD^l_L \big(\hat{u}^l_{t+i}~||~u_t^l~||~u_{t+n}^l~||~\textsf{up}.(u_{t+i}^{l+1})\big)
    \label{eqn:local}
    \vspace{-1mm}
\end{equation}

where $i = \{1,\ldots,m\}$, \textsf{up}. stands for upsampling and $u^l_{t+i}$ is the forecasted feature at level $l$. In principle, $\calD_L$ can decode both local and global motions. However, explicitly modelling global motions with $\calD_G$ is shown to be effective for the task at hand (see \Sref{sec:ablation}).
\begin{figure}[!t]
    \centering
    \includegraphics[width=1.0\linewidth,trim={20.3cm 23.9cm 33.1cm 7.8cm},clip]{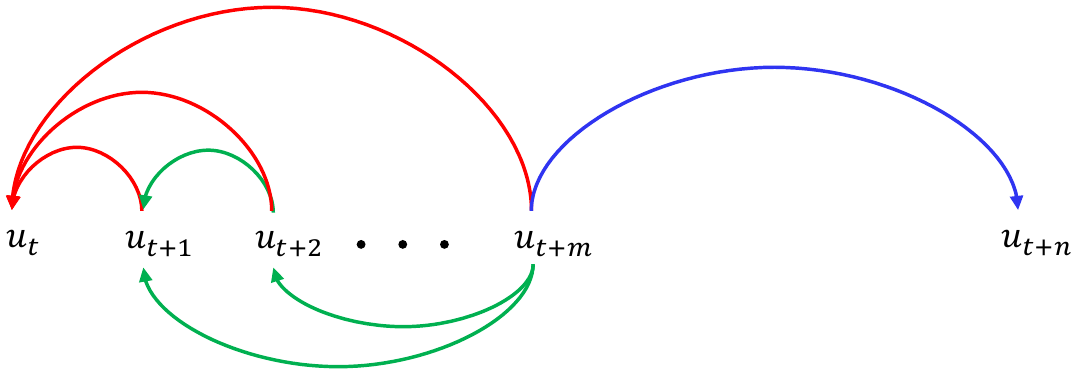}
    \vspace{-5.5mm}
    \caption{Optical flow estimation pattern when propagating to $m$ future counterparts.}
    \vspace{-6.3mm}
    \label{fig:flow_pattern}
\end{figure}
\vspace{-2.5mm}
\paragraph{Optical Flow Estimation.}  \textsf{M2FNet} learns to propagate features via motion supervision. For instance, to ensure that the forecasted feature $u_{t+i}$ can be reconstructed to $x_{t+i}$, we constrain the endpoint error between the flows $f_{x_{t+i} \rightarrow x_t}$ and $\hat{f}_{u_{t+i} \rightarrow u_t}$, which are computed between frames $\langle x_t,x_{t+i}\rangle$ and features $\langle u_t,u_{t+i}\rangle$, respectively. As the ground truth flow  $f_{x_{t+i} \rightarrow x_{t}}$ does not exist for real high-speed videos, we generate a pseudo-ground truth flow using pretrained state-of-the-art optical flow models~\cite{IMKDB17,Sun2018PWC-Net}. To estimate the flow $\hat{f}_{u_{t+i} \rightarrow u_t}$, we design a feature flow network (\textsf{FFNet}) which inputs two sets of features, \ie~$\big\{u_{t+i}^l\big\}_{l=1}^{k}$ and $\big\{u_{t}^l\big\}_{l=1}^{k}$, and regresses a flow in a coarse-to-fine manner. We use the architecture of the optical flow estimator in PWC-Net\cite{Sun2018PWC-Net} for \textsf{FFNet}. To predict the flow between the anchor feature $u_t$ and a forecasted feature $u_{t+i}$, we preform the following steps. First, at each level $l$, we  backwarp the second feature $u_{t}^l$ (to the first feature $u_{t+i}^l$) with $\times 2$ upsampled flow from previous level $l+1$ (see \Eref{eqn:warp}). A correlation layer \cite{fischer2015flownet,Sun2018PWC-Net,hui2018liteflownet} is then used to compute the cost volume between the first feature $u_{t+i}^l$ and the backwarped feature $w_{t+i}^l$. The first feature, the cost volume and the upscaled flow  are feed into \textsf{FFNet} to predict a flow as shown in \Eref{eqn:decode_flow}.
\begin{equation}
    w_{t+i}^l = \textsf{backwarp}\big(u_{t}^l, \textsf{up}.(\hat{f}^{l+1}_{{t+i}\rightarrow t})\big)
    \label{eqn:warp}
\vspace{-4mm}
\end{equation}
\begin{equation}
    \hat{f}_{{t+i}\rightarrow t}^l = \textsf{FFNet} \big(u_{t+i}^l \oplus \textsf{corr}.(u_{t+i}^l,w_{t+i}^l) \oplus \textsf{up}.(\hat{f}^{l+1}_{{t+i}\rightarrow t})\big)
    \label{eqn:decode_flow}
\end{equation}
\Fref{fig:flow_pattern} depicts the flow estimation pattern when propagating to $m$ future counterparts of the anchor feature $u_t$. We compute several optical flows to ensure that features are propagated by anticipating the complex motion between the input frames and not simply in a linear manner. Specifically, we estimate the flow between the anchor feature and each forecasted feature (shown in \textcolor{red} {red} in \Fref{fig:flow_pattern}) so that \textsf{M2FNet} learns to forecast features according to their proximity to the anchor feature \ie~$\calD_G$ and $\calD_L$ decode smaller motions for features close to the anchor feature and larger motions for those further away. We also estimate flow between the forecasted features themselves (depicted in \textcolor{green} {green}) in order to account for inter-frame motion between propagated features. To address any potential ambiguity in the direction of propagation, we compute the flow between the last forecasted feature and the reference feature (shown in \textcolor{blue} {blue}).
\vspace{-2.5mm}
\paragraph{Feature-to-Frame Decoding.} The forecasted features and optical flows are then used to decode frames. For this purpose, we design a frame decoder ($\calD_F$) which regresses frames from the corresponding forecasted features. When decoding the current frame, $\calD_F$ incorporates contextual and temporal information from the past frames via attention mechanism (see \Fref{fig:frame_synthesis}). This is accomplished by warping the past features into the current time step with the corresponding estimated optical flows and combining the warped features using attention weights as shown in \Eref{eqn:attend}. The attention vector ($\alpha$) is a learnable, one-dimensional weight parameter with elements initially set to 1. For better reconstruction of occluded regions in the predicted frames, $\calD_F$ also uses the anchor and reference features. Like the feature forecasting and flow estimation steps, frames are decoded in a coarse-to-fine manner. At each feature level $l$, $\calD_F$ inputs the forecasted feature ($u_{t+i}^l$), attended past features ($v_{t+i}^l$), encoded features ($u_t^l$ and $u_{t+n}^l$) and $\times 2$ upscaled frame predicted from previous level $l+1$ (see \Eref{eqn:decode_frame}). $\calD_F$ is composed of 3 densely connected convolutional layers each with kernel size $3 \times 3$ and stride $1$, where the last layer outputs a frame. 
\vspace{-1.2mm}
\begin{equation}
    {v}_{t+i}^l = \sum_{j=t}^{t+i-1}{\alpha_j . \textsf{backwarp}(u^l_j, \hat{f}^l_{{t+i}\rightarrow j})}
    \label{eqn:attend}
\end{equation}
\vspace{-3mm}
\begin{equation}
    \hat{x}^l_{t+i} = \calD^l_F\big(u_{t+i}^l~||~v_{t+i}^l~||~u_t^l~||~u_{t+n}^l~||~\textsf{up}.(\hat{x}^{l+1}_{t+i})\big)
    \label{eqn:decode_frame}
    \vspace{-0.5mm}
\end{equation}

\vspace{-2mm}
\subsection{Propagation-Interpolation Network (\textsf{\textbf{P-INet}})}
\vspace{-1mm}
The proposed propagation network (\textsf{PNet}) can be used either as a stand-alone model or a plug-in module with existing VFI works (see \Sref{sec:experiment}). However, we have experimentally observed few trade-offs. First, when the temporal gap between inputs is small, \textsf{PNet} gives a sub-optimal performance compared to the state-of-the-art VFI approaches. This is mainly because \textsf{PNet}, by design nature, propagates to future counterparts of the anchor frame by using the other input as a reference, \ie it relies more on one of the inputs by default. This leads to a performance trade-off since an interpolation method, intuitively speaking, should evenly rely on both inputs when there is a small gap between them. Second, as expected, the quality of frames eventually deteriorates as we propagate further away from the anchor frame. For completeness of the proposed approach, thereby alleviating the observed trade-offs, we adopt a light version of \textsf{SloMo} \cite{slomo} as an interpolation network (\textsf{INet}) along with \textsf{PNet} as our full model, \ie \textsf{P-INet}. The \textsf{SloMo} used in our work contains 50\% fewer weight parameters compared to the model used in \cite{slomo}. We train \textsf{P-INet} in an end-to-end manner by guiding it to propagate, interpolate or propagate and interpolate depending on the \textit{temporal gap} between inputs and the \textit{timestamp} of the intermediate frame to be predicted as summarized in \Aref{algo:train_algo}. 

We use a bidirectional propagation and interpolation scheme since it gives us the flexibility to experiment with long-range temporal gaps. The reliable time frame of propagation $\Delta t (n)$ is defined as $\min(\ceil*{(n-M)/2}, M)$, where $n > M$. In other words, \textsf{PNet} adaptively propagates until the temporal gap between the end propagated frames is less than or equal to $M$. As most VFI works conduct experiments at 30 fps by downsampling 240 fps videos ($\approx$ frame gap of 8) and because our approach uses 3 intervals, we set $M=8$ and $N=24$ during training. We experiment with up to 30 frame gaps during testing to analyze if our approach extends to even larger gaps (see \Sref{sec:experiment}). 

\setlength{\textfloatsep}{1.5mm}
\begin{algorithm}[!t]
\footnotesize
\LinesNotNumbered
\DontPrintSemicolon
\SetAlgoLined
\SetNoFillComment
 \textbf{Input}\ : $\langle x_{t},x_{t+n}\rangle$ \tcp*{$n$ is the frame gap} 
 \textbf{Output}\ : $\hat{x}_{t+i}$, where $1<i<n$

 Let $N$ be the maximum frame gap in the dataset, $M$ be the upper limit for small frame gap, and $\Delta t (n)$ be a reliable time frame of propagation which is dependent on $n$ \;
 \ForEach{input sample}{
  
  \eIf (\tcp*[h]{small gap}){$n \le M$}{
  $\hat{x}_{t+i} = $ \textsf{INet} $(x_{t},x_{t+n})$ \textbf{for} all $i$ \;
  }(\tcp*[h]{large gap ($M < n \le N$)}){
  \uIf(\tcp*[h]{propagate from $x_t$}){$i \le \Delta t (n)$}{
  $\hat{x}_{t+i} = \textsf{PNet}$ $(x_{t},x_{t+n})$\;
  }
  \uElseIf{$ \Delta t (n) < i < n-\Delta t (n)$}{
  \tcp*[h]{propagate and interpolate}\
    $\hat{x}_{t + \Delta t (n)} = \textsf{PNet}$ $(x_{t},x_{t+n})$\;
    $\hat{x}_{t+ n - \Delta t (n)} = \textsf{PNet}$ $(x_{t+n},x_{t})$\;
     $\hat{x}_{t+i} \ = $ \textsf{INet} $(\hat{x}_{t + \Delta t (n)},\hat{x}_{t+ n - \Delta t (n)})$\;
  }
  \uElse(\tcp*[h]{propagate from $x_{t+n}$}){
  $\hat{x}_{t+i} = \textsf{PNet}$  $(x_{t+n},x_{t})$\;
  }
  }
 }
 \caption{\footnotesize{Training strategy for the \textsf{P-INet}}}
 \label{algo:train_algo}
\end{algorithm}
\vspace{-0.4mm}
\subsection{Loss Functions}
\vspace{-0.5mm}
We train our network in an end-to-end manner by jointly optimizing the estimated flows, propagated frames and interpolated frames. To train \textsf{M2FNet}, we compute the endpoint error between the estimated flows and pseudo-ground truth flows across different levels as shown in \Eref{eqn:loss_flow}. To propagate to $m$ features, a total of $\frac{m}{2}(m+1)+1$ flows are estimated. We also investigate training our network by selectively optimizing some of the flows (see \Sref{sec:ablation}).
\vspace{-1mm}
\begin{equation}
    \calL_\textsf{M2FNet} = \sum_{i=1}^{\frac{m}{2}(m+1)+1}\sum_{l=1}^{k}\omega_1^l\big|f_i^l-\hat{f}_i^l\big|_{2}
    \label{eqn:loss_flow}
    \vspace{-1.5mm}
\end{equation}
where $\omega_1^l$ is a flow loss weight coefficient at level $l$. For sharp frame decoding, we train \textsf{PNet} with the multi-scale $\ell_1$ photometric loss. We also use gradient difference loss \cite{mathieu2015deep} ($\calL_\textsf{GDL}$) between the predicted frames and their ground truth to mitigate blurry predictions (see \Eref{eqn:loss_frame}). 
\vspace{-1.5mm}
\begin{equation}
    \calL_\textsf{PNet} = \sum_{i=1}^{m}\sum_{l=1}^{k}\omega_2^l\big|x_i^l-\hat{x}_i^l\big|_{1} + \calL_\textsf{GDL} (x_i, \hat{x}_i) 
    \label{eqn:loss_frame}
    \vspace{-1.35mm}
\end{equation}
where $\omega_2^l$ is a frame loss weight coefficient at level $l$. We use the training loss of \textsf{SloMo} discussed in Section 3.3 of \cite{slomo} to train \textsf{INet}. We refer the reader to \cite{slomo} for details. The total training loss for \textsf{P-INet} is defined as weighted sum of all losses as shown in \Eref{eqn:loss_total}.
\vspace{-2mm}
\begin{equation}
    \calL_\textsf{total} = \lambda_1\calL_\textsf{M2FNet} + \lambda_2\calL_\textsf{PNet} + \lambda_3\calL_\textsf{INet}
    \label{eqn:loss_total}
    \vspace{-2.5mm}
\end{equation}

\vspace{-1.5mm}
\section{Experiment}
\label{sec:experiment}
\vspace{-0.5mm}
\begin{table*}[!t]
\caption{Quantitative comparison at different fps. The numbers in \textcolor{red}{red} and \textcolor{blue}{blue} represent the best and second best results, respectively.}
\label{tbl:temp_vfi}
\vspace{-2.5mm}
\begin{adjustbox}{width=\linewidth}
\setlength{\tabcolsep}{4.5pt}
\renewcommand{\arraystretch}{1.035}
\begin{tabular}{l|cccccc|cccccc|cccccc}
\toprule
& \multicolumn{6}{c|}{\textsf{Adobe240} \cite{su2017deep}} & \multicolumn{6}{c|}{\textsf{GOPRO}~\cite{Nah_2017_CVPR}} & \multicolumn{6}{c}{\textsf{NfS}~\cite{galoogahi2017need}}  \\ \cmidrule(lr){2-7} \cmidrule{8-13} \cmidrule(lr){14-19}
Method & \multicolumn{2}{c}{30 fps} & \multicolumn{2}{c}{15 fps} & \multicolumn{2}{c|}{8 fps} & \multicolumn{2}{c}{30 fps} & \multicolumn{2}{c}{15 fps} & \multicolumn{2}{c|}{8 fps} & \multicolumn{2}{c}{30 fps} & \multicolumn{2}{c}{15 fps} & \multicolumn{2}{c}{8 fps}  \\ \cmidrule(lr){2-3} \cmidrule(lr){4-5} \cmidrule(lr){6-7} \cmidrule(lr){8-9} \cmidrule(lr){10-11} \cmidrule(lr){12-13}\cmidrule(lr){14-15}\cmidrule(lr){16-17}\cmidrule(lr){18-19} 
& PSNR & SSIM & PSNR & SSIM & PSNR & SSIM & PSNR & SSIM & PSNR & SSIM & PSNR & SSIM & PSNR & SSIM & PSNR & SSIM & PSNR & SSIM \\ \midrule
\textsf{SepConv}~\cite{niklaus2017video2}&29.91 &0.915&23.94&0.811& 19.88&0.707&28.64&0.871&23.23&0.694&19.74&0.560&31.84&0.915&26.73&0.811&23.00&0.707 \\ 
\textsf{SloMo}~\cite{slomo} &30.03&0.917&24.30&0.818&20.17&0.717&29.03&0.917&23.58&0.818&19.99&0.718&31.83&0.917&26.95&0.818&23.19&0.717  \\
\textsf{DAIN}~\cite{dain2019} &\textcolor{red}{30.53}&\textcolor{red}{0.924}&\textcolor{blue}{24.39}&\textcolor{blue}{0.824}&\textcolor{blue}{20.21}&\textcolor{blue}{0.721}&\textcolor{blue}{29.25}&\textcolor{red}{0.924}&\textcolor{blue}{23.63}&\textcolor{blue}{0.824}&\textcolor{blue}{20.18}&0.721&\textcolor{red}{32.46}&\textcolor{red}{0.924}&\textcolor{blue}{27.19}&\textcolor{blue}{0.824}&\textcolor{blue}{23.36}&0.720 \\ 
\textsf{AdaCoF}~\cite{Lee_2020_CVPR}&30.14&0.896&24.11&0.741&20.07&0.567&29.05&0.876&23.49&0.701&19.89&0.571&32.28&0.919&27.05&0.819&23.23&0.719  \\ 
\textsf{FeFlow}~\cite{gui2020featureflow} &\textcolor{blue}{30.48}&{0.902}&24.19&0.737&20.04&0.576&\textcolor{red}{29.30}&\textcolor{blue}{0.921}&23.51&0.822&19.82&\textcolor{blue}{0.724}&\textcolor{blue}{32.42}&\textcolor{blue}{0.921}&27.05&0.822&23.16&\textcolor{blue}{0.724} \\ \midrule
\textsf{INet} &30.30&0.920&24.21&0.819&20.12&0.718&29.17&0.919&23.59 &0.821&20.04&0.722&32.03&0.920&26.99&0.822&23.27&0.721  \\ 
\textbf{\textsf{P-INet}} &30.30&\textcolor{blue}{0.920}&\textcolor{red}{27.10}&\textcolor{red}{0.890}&\textcolor{red}{24.00}&\textcolor{red}{0.810}&29.17&0.919&\textcolor{red}{26.45}&\textcolor{red}{0.879}&\textcolor{red}{23.90}&\textcolor{red}{0.804}&32.03&0.920&\textcolor{red}{28.98}&\textcolor{red}{0.874}&\textcolor{red}{26.23}&\textcolor{red}{0.798} \\ \bottomrule
\end{tabular}
\end{adjustbox}
\vspace{-2.8mm}
\end{table*}
\begin{figure*}[!t]
\begin{center}
\setlength{\tabcolsep}{0.3pt}
\renewcommand{\arraystretch}{0.2}
\resizebox{1.0\linewidth}{!}{%
\begin{tabular}{cccccc}
        \tiny{\textsf{Inputs Overlay}} & \tiny{\textsf{DAIN} \cite{dain2019}} & \tiny{\textsf{AdaCoF} \cite{Lee_2020_CVPR}} & \tiny{\textsf{FeFlow} \cite{gui2020featureflow}} & \tiny{\textsf{\textbf{P-INet}}} & \tiny{\textsf{GT}} \\
          \includegraphics[width=0.1\linewidth]{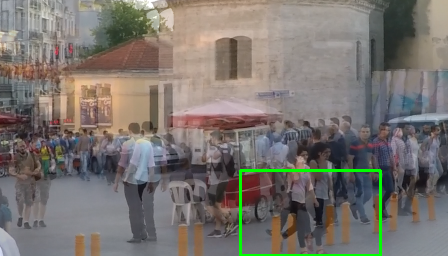}&
         \includegraphics[width=0.1\linewidth]{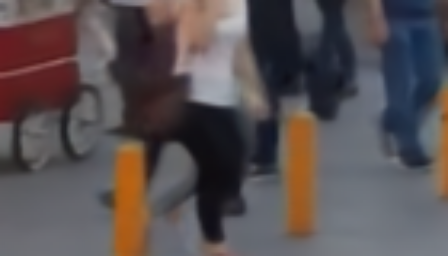}&
         \includegraphics[width=0.1\linewidth]{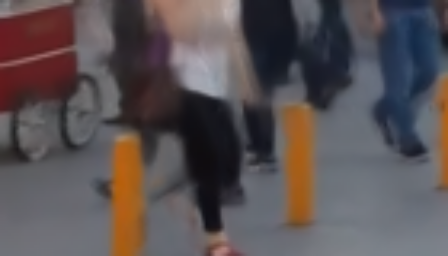}&
         \includegraphics[width=0.1\linewidth]{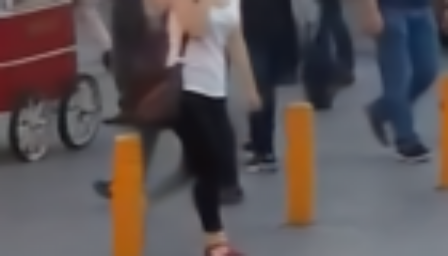}&
         \includegraphics[width=0.1\linewidth]{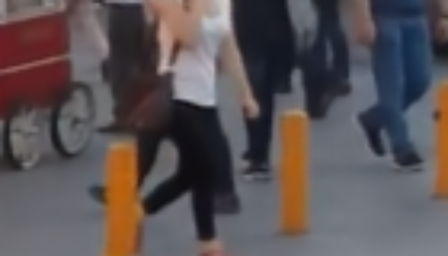}&
         \includegraphics[width=0.1\linewidth]{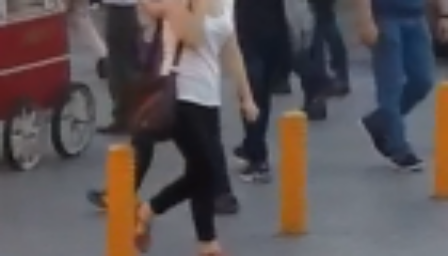}\\
         \includegraphics[width=0.1\linewidth]{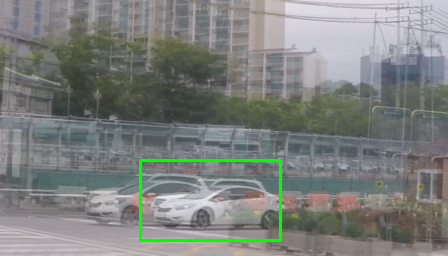}&
         \includegraphics[width=0.1\linewidth]{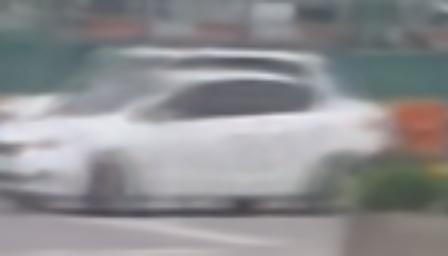}&
         \includegraphics[width=0.1\linewidth]{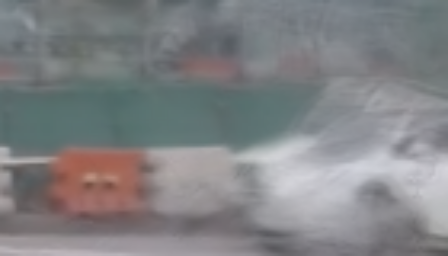}&
         \includegraphics[width=0.1\linewidth]{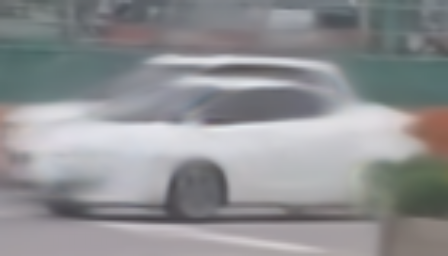}&
         \includegraphics[width=0.1\linewidth]{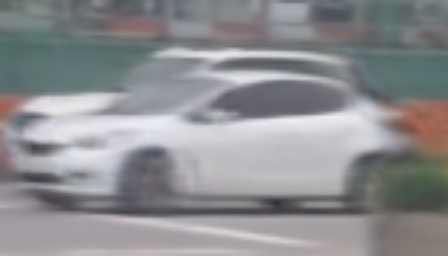}&
         \includegraphics[width=0.1\linewidth]{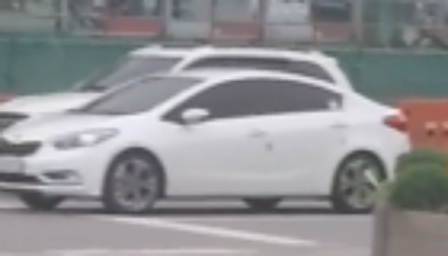}\\
\end{tabular}}
\end{center}
\vspace{-6mm}
\caption{Qualitative comparison of our method and state-of-the-art VFI approaches for inputs with large temporal gap.}
\label{tbl:qual_compare}
\vspace{-5.5mm}
\end{figure*}
\paragraph{Datasets.} Most existing VFI works use Vimeo-90K \cite{xue2019video} dataset which has 51312 triplets, where each triplet contains 3 consecutive video frames. However, as this dataset is not applicable to train a network for long-term VFI, we generate a dataset by sampling frames at different fps from high-speed video datasets. For this purpose, we use \textsf{Adobe240} \cite{su2017deep}, \textsf{GOPRO} \cite{Nah_2017_CVPR} and Need-for-Speed (\textsf{NfS}) \cite{galoogahi2017need} datasets, which contain 133, 33 and 100 videos, respectively. These datasets provide 240 fps videos which capture diverse combination of camera and object motions in real-world scenarios, and thus are suitable for the task at hand. A majority of the videos, however, have less than 1000 frames which makes it challenging to extract enough training samples with large temporal gaps. Hence, instead of training separately on each dataset, we used a total of 176 videos (103 from \textsf{Adobe240}, 3 from \textsf{GOPRO} and 70 from \textsf{NfS}) for training. The remaining 90 videos (30 from each dataset) are used for testing. We prepare train and test sets by extracting samples with variable length ranging from 9 to 31 consecutive frames in a video. In other words, we sample video clips at different frame rates in the range of approximately 30 fps to 8 fps, respectively. Following \cite{xue2019video}, we resize each frame in the dataset to a resolution of $448 \times 256$ to suppress noise and create consistency in size across videos.
\vspace{-6mm}
\paragraph{Implementation Details.}
We implement our network in PyTorch \cite{paszke2019pytorch} and optimize it using Adam \cite{KingmaB14} with parameters $\beta_1$ , $\beta_2$ and \textit{weight decay} fixed to $0.9$, $0.999$ and $4e-4$, respectively. The loss weight coefficients are set to $\omega^5 = 0.08$, $\omega^4 = 0.04$, $\omega^3 = 0.02$, $\omega^2 = 0.01$ and $\omega^1 = 0.005$ from the lowest to the highest resolution, respectively, for both $\calL_\textsf{M2FNet}$ and $\calL_\textsf{PNet}$. We train \textsf{P-INet} for 200 epochs with the learning rate initially set to $\lambda=1e-4$ and gradually decayed by half at 100, 150 and 175 epochs. For the first 40 epochs, we only train the \textsf{M2FNet} by setting $\lambda_1 = 1$, $\lambda_2 = 0$  and  $\lambda_3 = 0$ to facilitate motion estimation and feature propagation. For the remaining epochs, we fix $\lambda_1$, $\lambda_2$ and $\lambda_3$ to 1.  We use a mini-batch size of 4 and randomly crop image patches of size $256 \times 256$ during training. The pseudo-ground truth optical flows for supervising \textsf{M2FNet} are computed on-the-fly using FlowNet 2 \cite{IMKDB17}.
\vspace{-0.5mm}
\subsection{Experimental Results}
\vspace{-0.6mm}
In this section, we comprehensively analyze our work and several state-of-the-art VFI approaches for which open source implementations are available. These include \textsf{SepConv} \cite{niklaus2017video2}, \textsf{SloMo}~\cite{slomo}, \textsf{DAIN}~\cite{dain2019}, \textsf{AdaCoF}~\cite{Lee_2020_CVPR} and \textsf{FeFlow}~\cite{gui2020featureflow}. For fair comparison, we retrain these models using our training set by following their official code. We deploy a multi-frame interpolation training scheme for \textsf{P-INet}, \textsf{SloMo}~\cite{slomo} and \textsf{DAIN}~\cite{dain2019} as it is possible while we use single-frame interpolation scheme for others. For quantitative evaluation, we use PSNR and SSIM metrics.
\vspace{-3.1mm}
\paragraph{Temporally Robust VFI.}
Here, we analyze the robustness of different VFI models for input sequences with different temporal gaps. In \Tref{tbl:temp_vfi}, we compare our approach and state-of-the-art VFI methods on single frame interpolation of test videos sampled at 3 different frame rates: 30 fps, 15 fps and 8 fps. As can be inferred from \Tref{tbl:temp_vfi}, \textsf{P-INet} performs competitively for smaller temporal gaps and significantly better than SOTA approaches for larger temporal gaps. For instance, our approach outperforms the second best method, ~\ie \textsf{DAIN}\cite{dain2019}, by an average margin of of 2.44 dB and 3.51 dB at 15 fps and 8 fps, respectively. Moreover, the performance gap for \textsf{DAIN} between 30 fps and 8 fps is 9.50 dB on average. By contrast, the performance gap for our model is 5.79 dB. This shows the effectiveness of our approach for low-frame-rate videos. It can also be noticed from \Tref{tbl:temp_vfi} that the joint training of \textsf{PNet} and \textsf{INet} is beneficial even for smaller frame gaps. For instance, \textsf{INet} outperforms \textsf{SloMo} \cite{slomo} by an average margin of 0.2 dB at 30 fps. In \Fref{tbl:qual_compare}, we qualitatively compare the frames interpolated by our method and SOTA VFI approaches for input samples with large temporal gap. As can be seen from the figure, our approach interpolates sharper images with clearer contents compared to other VFI approaches.
\vspace{-3.65mm}
 \begin{figure*}[!t]
    \centering
    \includegraphics[width=1.0\linewidth,trim={0.3cm 0.35cm 0.3cm 0.1cm},clip]{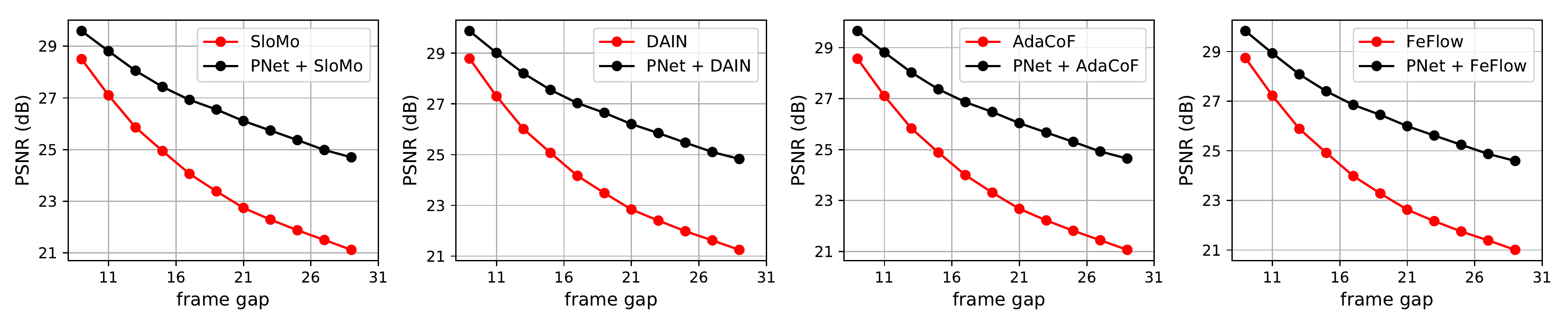}
    \vspace{-6.5mm}
    \caption{Quantitative analysis of \textsf{PNet} with state-of-the-art VFI approaches.}
    \label{fig:fpnet_interp}
    \vspace{-2.5mm}
\end{figure*}

\begin{figure*}[!t]
\begin{center}
\setlength{\tabcolsep}{0.3pt}
\renewcommand{\arraystretch}{0.2}
\resizebox{1.0\linewidth}{!}{%
\begin{tabular}{ccccccc}
        \tiny{\textsf{Inputs Overlay}} & \tiny{\textsf{FeFlow} \cite{gui2020featureflow}} & \tiny{\textsf{\textbf{PNet}} + \textsf{FeFlow}\cite{gui2020featureflow}} & \tiny{\textsf{DAIN} \cite{dain2019}} &
        \tiny{\textsf{\textbf{PNet}} + \textsf{DAIN} \cite{dain2019}} & \tiny{\textsf{GT}} \\
        \includegraphics[width=0.1\linewidth]{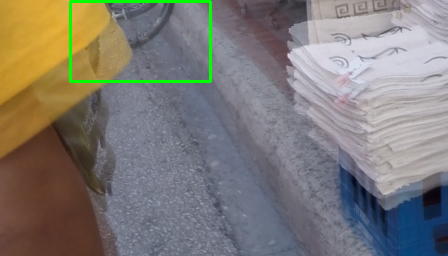}&
        \includegraphics[width=0.1\linewidth]{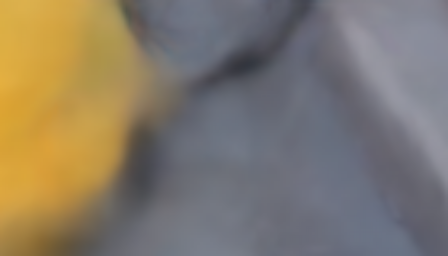}&
        \includegraphics[width=0.1\linewidth]{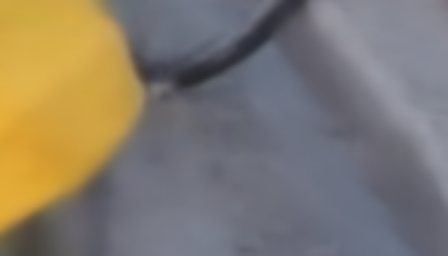}&
        \includegraphics[width=0.1\linewidth]{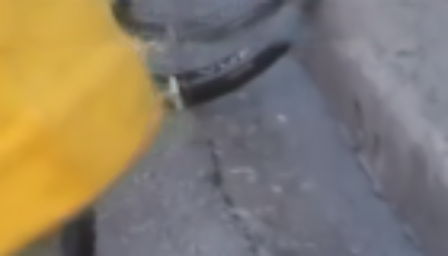}&
        \includegraphics[width=0.1\linewidth]{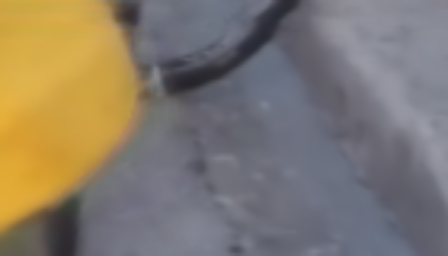} &
        \includegraphics[width=0.1\linewidth]{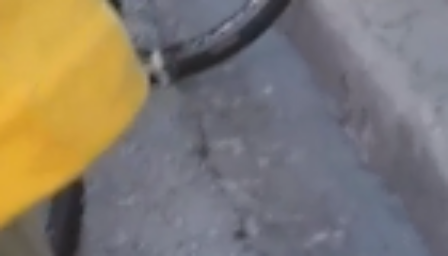}\\

        \includegraphics[width=0.1\linewidth]{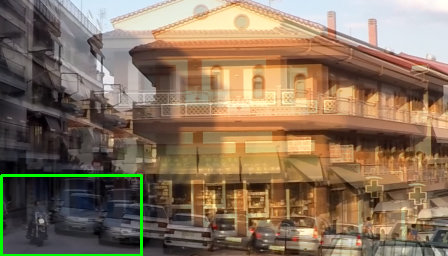}&
        \includegraphics[width=0.1\linewidth]{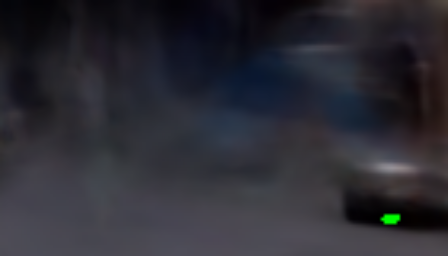}&
          \includegraphics[width=0.1\linewidth]{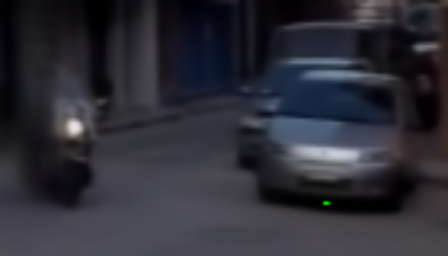}&
          \includegraphics[width=0.1\linewidth]{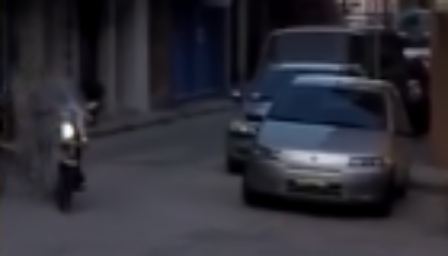}&
          \includegraphics[width=0.1\linewidth]{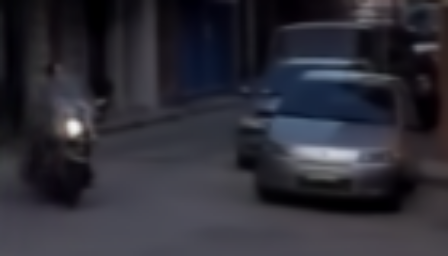}&
         \includegraphics[width=0.1\linewidth]{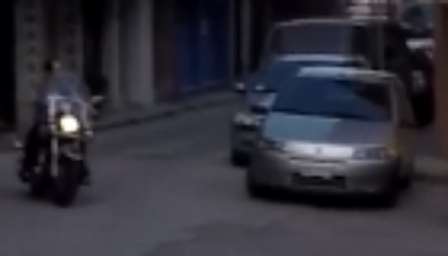}
\end{tabular}}
\end{center}
\vspace{-6.5mm}
\caption{Qualitative analysis of \textsf{PNet} cascaded with state-of-the-art VFI approaches.}
\label{tbl:qual_fpnet}
\vspace{-2.5mm}
\end{figure*}
\begin{figure*}[!t]
    \centering
    \includegraphics[width=1.0\linewidth,trim={0.3cm 0.15cm 0.25cm 0.25cm},clip]{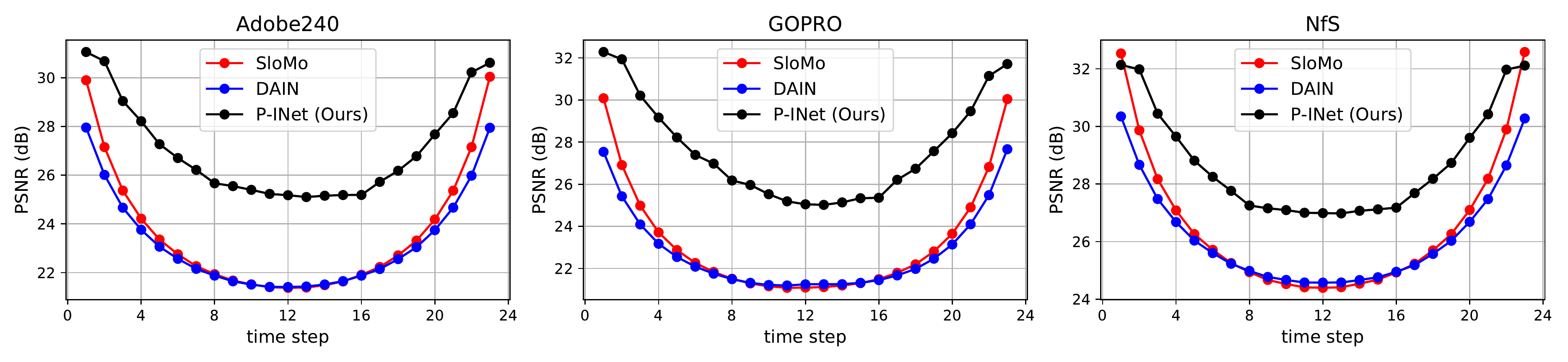}
    \vspace{-6.3mm}
    \caption{Quantitative analysis of intermediate frames at different time steps for long-term VFI.}
    \vspace{-5.3mm}
    \label{fig:temp_smooth}
\end{figure*}
\paragraph{\textsf{PNet} with VFI Methods.}
To highlight the versatility of the proposed \textsf{PNet} for long-term VFI, we couple \textsf{PNet} with VFI approaches and perform intermediate frame interpolation for input sequences with relatively large frame gap ranging from 11 to 30. Following the procedure in \Aref{algo:train_algo}, we first propagate bidirectionally using \textsf{PNet} from a pretrained \textsf{P-INet}. Then, we interpolate an intermediate frame between the propagated frames using state-of-the-art VFI methods~\cite{slomo,dai2008motion,Lee_2020_CVPR,gui2020featureflow}. The averaged results on the 3 datasets are plotted in \Fref{fig:fpnet_interp}. As can be inferred from the figure, the cascade models consistently outperform their vanilla baseline by a notable margin. The qualitative analysis in \Fref{tbl:qual_fpnet} also shows that, when the temporal distance between the inputs is large, incorporating \textsf{PNet} results in interpolated frames with more \textit{accurate} contents compared to directly using SOTA VFI approaches. The static regions in \Fref{tbl:qual_fpnet} appear slightly less sharp for cascade models most likely because the interpolation model uses the output of \textsf{PNet} as inputs rather than the raw input frames. 
\vspace{-4mm}
\paragraph{Long-term Multi-Frame Interpolation.}
Beyond evaluating the robustness of VFI approaches at different frame rates, we analyze the quality of the intermediate frames interpolated during a direct \textit{very low} fps $\rightarrow$ \textit{very high} fps upsampling. We perform 10 fps $\rightarrow$ 240 fps up-conversion in a single pass and measure the quality of the interpolated frames at each time step. In \Fref{fig:temp_smooth}, we compare our approach with \textsf{SloMo} \cite{slomo} and \textsf{DAIN} \cite{dain2019} since they are also capable of multi-frame interpolation. As expected, performance generally decreases as we move to the middle time step from both sides. However, it can be noticed from \Fref{fig:temp_smooth} that there is a rapid performance drop for \textsf{SloMo} and \textsf{DAIN} compared to \textsf{P-INet}. For instance, the average performance range, \ie the difference between largest and smallest PSNR values averaged over the 3 datasets, for \textsf{SloMo} is 8.62 dB. By contrast, the average performance range for \textsf{P-INet} is 6.11 dB. Instead of interpolating frames based on pre-computed motion that will likely be inaccurate due to large motion, our model adapts to propagate and interpolate frames, which explains the significant performance gain achieved over state-of-the-art approaches particularly for central time steps.
\vspace{-4.5mm}
\begin{figure*}[!t]
\begin{center}
\setlength{\tabcolsep}{0.3pt}
\renewcommand{\arraystretch}{0.5}
\resizebox{1.0\linewidth}{!}{%
\begin{tabular}{cccccc}
        \tiny{\textsf{Input} ($x_t$) } & \tiny{$\hat{f}$ (\textsf{Ours})} & \tiny{$f$ (\textsf{p-GT})} & \tiny{\textsf{Input} ($x_t$) } & \tiny{$\hat{f}$ (\textsf{Ours})} & \tiny{$f$ (\textsf{p-GT})} \\
         \includegraphics[width=0.1\linewidth]{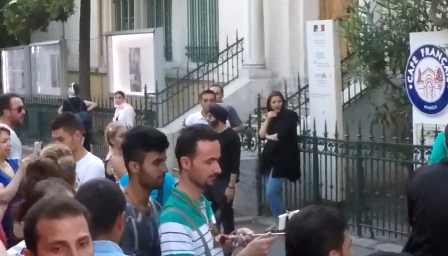}&
        \includegraphics[width=0.1\linewidth]{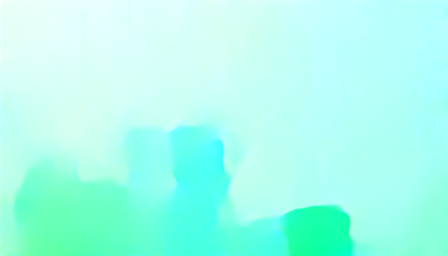}&
         \includegraphics[width=0.1\linewidth]{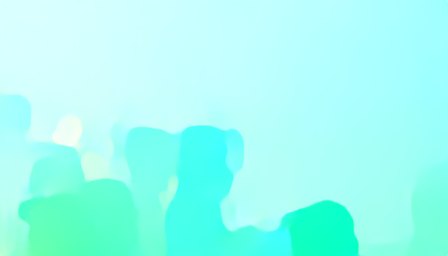}&
         
        \includegraphics[width=0.1\linewidth]{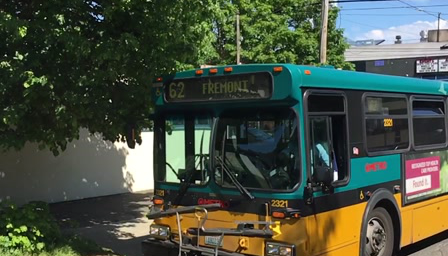}&
        \includegraphics[width=0.1\linewidth]{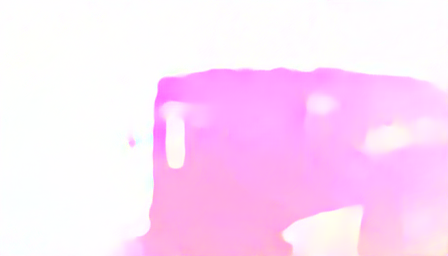}&
         \includegraphics[width=0.1\linewidth]{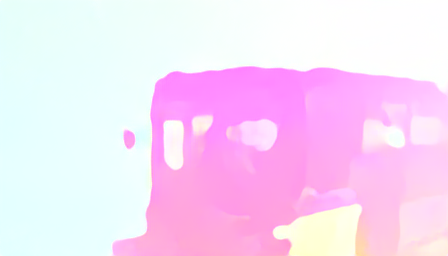}
\end{tabular}}
\end{center}
\vspace{-6.3mm}
\caption{Qualitative analysis of the estimated optical flows between features in comparison with the pseudo-ground truth (\textsf{p-GT}) flows.}
\label{tbl:qual_compare_flow1}
\vspace{-5mm}
\end{figure*}
\begin{figure}[!t]
    \centering
    \includegraphics[width=1\linewidth,trim={0.5cm 0.4cm 0.7cm 0.45cm},clip]{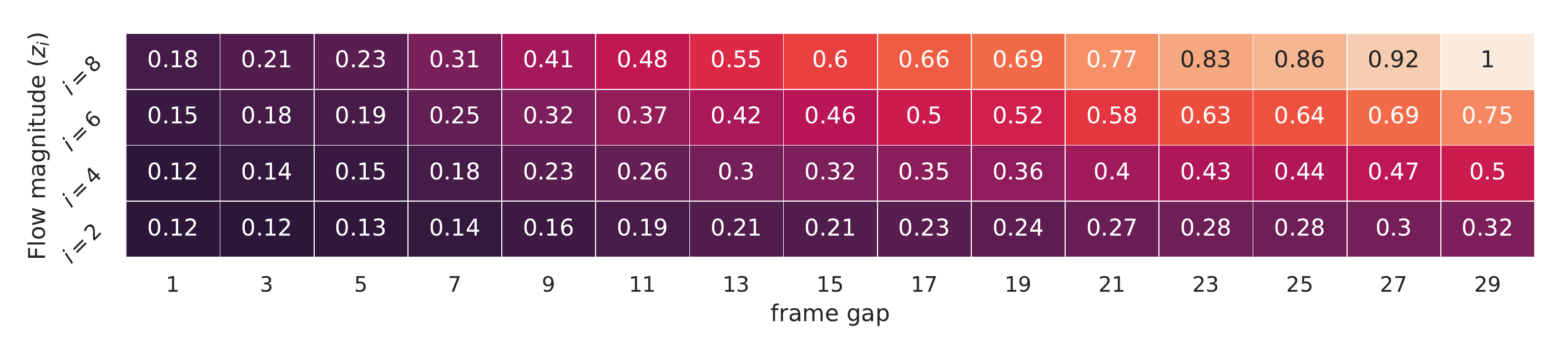}
     \vspace{-6mm}
    \caption{Quantitative analysis of the estimated flows.}
    \label{fig:heatmap}
\end{figure}
\paragraph{Optical Flow.} \Fref{tbl:qual_compare_flow1} depicts the feature flows estimated by our network in comparison with the corresponding pseudo-ground truth (\textsf{p-GT}) flows. As can be seen from \Fref{tbl:qual_compare_flow1}, our model reasonably anticipates the accurate motion to propagate features. To further confirm if $\calD_G$ and $\calD_L$ in the \textsf{M2FNet} properly learned to decode motions for feature propagation, we quantitatively analyze the optical flows estimated between the \textit{anchor} and the \textit{forecasted} features. To purely evaluate the magnitude of motion, we compute the sum of the absolute value of the estimated flows. In \Fref{fig:heatmap}, we plot a heat map of the magnitude of the estimated flows (rescaled between 0 \& 1) for different temporal gaps, where $z_i = \frac{1}{2}(\sum{|\hat{f}_{{t+i}\rightarrow t}|}+\sum{|\hat{f}_{{t+n-i}\rightarrow{t+n}}|})$. We can infer two key things from \Fref{fig:heatmap}. First, the proximity of the forecasted feature to the anchor is directly related to the magnitude of the estimated flow, \ie  $\calD_G$ and $\calD_L$ decode  smaller motions for closer features and larger motion for those that are far. Second, \textsf{M2FNet} is implicitly aware of the relative temporal distance between inputs, \ie the magnitude of the forecasted flows increases for increasing frame gap.
\vspace{-2.3mm}
\section{Ablation Studies}
\vspace{-1.5mm}
\label{sec:ablation}
Here, we present ablation experiments on different components of \textsf{P-INet}. We evaluate the quality of all \textit{propagated} frames during long-term VFI (10 fps $\rightarrow$ 240 fps) for \textsf{Adobe240} \cite{su2017deep} and \textsf{GOPRO}\cite{Nah_2017_CVPR} test videos (see \Tref{tbl:ablation_flow}).
\vspace{-9mm}
\paragraph{Loss Functions.}
To highlight the importance of using optical flow as a guidance for feature propagation, we forecast features without estimating flows and directly regress frames from the respective forecasted features, \ie \textsf{P-INet} is trained without $\calL_\textsf{\textsf{M2FNet}}$. A network trained without motion supervision performed significantly worse compared to a model trained with motion supervision. We also confirmed the contribution of the different groups of flows estimated in \Sref{sec:methodology}. It can be inferred from \Tref{tbl:ablation_flow} that estimating optical flows between the forecasted features is crucial as a network trained without inter-frame motion supervision (shown in \textcolor{green}{green} in \Fref{fig:flow_pattern}) gives a subpar performance. Moreover, we studied the importance of addressing potential directional ambiguity by constraining the optical flow between the end propagated feature and the reference feature (shown in \textcolor{blue}{blue} in \Fref{fig:flow_pattern}). As can be seen from \Tref{tbl:ablation_flow}, training a network without direction supervision results in a performance decrease of 0.48 dB. We analyze the benefit of the gradient difference loss \cite{mathieu2015deep} ($\calL_\textsf{GDL}$) in mitigating blurry frame predictions. It can be noticed from \Tref{tbl:ablation_flow} that training our model with $\calL_\textsf{GDL}$ improves performance by an average margin of 0.65 dB. 
\vspace{-3.5mm}
\paragraph{\textsf{M2FNet}.} We examine the importance of global ($\calD_G$) and local ($\calD_L$) motion decoders in \textsf{M2FNet}. First, we only use $\calD_G$ for decoding motion. This resulted in a subpar network performance as $\calD_G$ is limited to anticipating only global motion at a feature-level and the local motion apparent in the test videos can not be effectively forecasted. In theory, $\calD_L$ can decode both local and global motions (without the need to explicitly model global motions with $\calD_G$) as CNNs are effective in motion estimation tasks \cite{Sun2018PWC-Net,IMKDB17}. This is also empirically evident as a network trained only using $\calD_L$ gives a competitive performance. However, using $\calD_G$ to forecast global motions proved to give a considerable performance boost of 0.92 dB.
\vspace{-4mm}
\paragraph{Frame Decoding.} We study the importance of incorporating features of past frames when decoding the current frame in \textsf{PNet}. As can be inferred from \Tref{tbl:ablation_flow}, only attending to the \textit{anchor} feature (only warping $u_t$ in \Eref{eqn:attend}) when synthesizing frames gives a notably lower performance compared to attending all past features. Moreover, not attending to any past feature (excluding $v_{t+i}$ from \Eref{eqn:decode_frame}) during frame decoding performs significantly worse.
\begin{table}[!t]
\begin{center}
\caption{Ablation experiments}
\vspace{-2.9mm}
\label{tbl:ablation_flow}
\setlength{\tabcolsep}{5pt}
\renewcommand{\arraystretch}{0.94}
\begin{adjustbox}{width=\linewidth}
\begin{tabular}{l|cccc}
\toprule
& \multicolumn{2}{c}{\textsf{Adobe240}\cite{su2017deep}} & \multicolumn{2}{c}{\textsf{GOPRO}\cite{Nah_2017_CVPR}}  \\   \cmidrule (lr){2-3} \cmidrule (lr){4-5} 
\textbf{Loss Functions} & PSNR & SSIM & PSNR & SSIM  \\ \midrule
w/o $\calL_\textsf{\textsf{M2FNet}}$ &25.09&0.730&25.16&0.728 \\
w/o inter-frame motion &25.81&0.776&26.11&0.776 \\
w/o direction supervision &27.13&0.801&27.83&0.806\\
w/o $\calL_\textsf{\textsf{GDL}}$ & 26.97  & 0.801 & 27.84 & 0.813\\ \midrule
\textbf{\textsf{M2FNet}} & & & & \\
w/o  $\calD_L$  &25.13&0.734&25.71&0.760 \\
w/o  $\calD_G$  &26.96&0.801&27.34&0.811\\ \midrule
\textbf{Frame Decoding} & & & & \\
only warping  $u_t$ in \Eref{eqn:attend} & 26.82&0.793 & 27.41 & 0.812 \\
excluding $v_{t+i}$ from \Eref{eqn:decode_frame} & 26.03 &0.781&26.57& 0.789\\ \midrule
\textbf{\textsf{P-INet}}  & $\textbf{27.70}$ & $\textbf{0.816}$ & $\textbf{28.43}$ & $\textbf{0.843}$ \\ \bottomrule
\end{tabular}
\end{adjustbox}
\end{center}
\vspace{-3.5mm}
\end{table}
\vspace{-1.95mm}
\section{Conclusion}
\vspace{-1mm}
Our work introduces a temporally robust VFI framework by adopting a feature propagation approach. The proposed motion supervision tailors the network for the task at hand as it enforces features to be propagated according to the motion between inputs irrespective of their contents. The adaptive cascading of \textsf{PNet} with a simple interpolation backbone has significantly improved the interpolation quality for low frame rate videos as briefly analyzed in \Sref{sec:experiment}. 
\vspace{-3mm}
\paragraph{Limitations.} The multi-scale approach along with aggregated motion estimation significantly increases the time complexity of our model. For instance, during 10 fps $\rightarrow$ 240 fps up-conversion given an input pair of size $448 \times 256$, \textsf{SloMo}~\cite{slomo} takes 0.32 secs while \textsf{P-INet} takes 3.37 secs. We experimentally observed failure cases when there is a fast-moving \textit{small} object in the foreground of a scene with a relatively large, dynamic background. In this scenario, \textsf{PNet} fails to detect and anticipate the motion of such objects, and instead imitates the input feature during propagation. This results in temporal jittering artifact in the interpolated video. Improving this limitation using a detection module \cite{yuan2019zoom} or an attention mechanism \cite{choi2020channel} would be an interesting future direction.

{\small
\bibliographystyle{ieee_fullname}
\bibliography{egbib}
}

\end{document}